\documentclass{amsart}

\usepackage{amsmath}
\usepackage{amsthm}
\usepackage{amsfonts}
\usepackage{amssymb}
\usepackage{graphicx}
\usepackage{xcolor}
\usepackage{hyperref}
\usepackage{subcaption}
\usepackage{etoolbox}
\usepackage[foot]{amsaddr}
\usepackage{etoolbox}
\usepackage{multicol}
\usepackage{multirow}
\usepackage{tcolorbox}
\usepackage{datetime}
    \newdateformat{monthyeardate}{%
  \monthname[\THEMONTH], \THEYEAR}
\usepackage{setspace}
\usepackage{longtable}
\usepackage[authoryear,compress]{natbib}
\usepackage{tikz}

\hypersetup{
    colorlinks,
    linkcolor={red!50!black},
    citecolor={blue!50!black},
    urlcolor={blue!80!black}
}

\makeatletter
    \newtheoremstyle{indented}
        {12pt}
        {12pt}
        {\addtolength{\@totalleftmargin}{3.5em}
        \addtolength{\linewidth}{-3.5em}
        \parshape 1 3.5em \linewidth}
        {}
        {\bfseries}
        {:}
            {.5em}
        {}
\makeatother

\makeatletter
    \newtheoremstyle{indentedProp}
        {12pt}
        {12pt}
        {\addtolength{\@totalleftmargin}{3.5em}
        \addtolength{\linewidth}{-3.5em}
        \parshape 1 3.5em \linewidth}
        {}
        {\bfseries}
        {:}
            {0.5em}
        {}
\makeatother

\theoremstyle{indented}
    
\theoremstyle{indentedProp}
    
\theoremstyle{indented}
    
\theoremstyle{indented}
    
\theoremstyle{indented}
    
\theoremstyle{indented}
    
\theoremstyle{indented}

\makeatletter
    \patchcmd{\NAT@test}{\else \NAT@nm}{\else \NAT@nmfmt{\NAT@nm}}{}{}
    \DeclareRobustCommand\citepos
        {\begingroup
        \let\NAT@nmfmt\NAT@posfmt
        \NAT@swafalse\let\NAT@ctype\z@\NAT@partrue
        \@ifstar{\NAT@fulltrue\NAT@citetp}{\NAT@fullfalse\NAT@citetp}}
    \let\NAT@orig@nmfmt\NAT@nmfmt
    \def\NAT@posfmt#1{\NAT@orig@nmfmt{#1's}}
\makeatother

\title[The Linguistic Blind Spot of Value-Aligned Agency]{The Linguistic Blind Spot of Value-Aligned Agency, Natural and Artificial}

\author{Travis LaCroix}

\address{Department of Philosophy \\ Dalhousie University}

\email{tlacroix@dal.ca}

\date{Unpublished draft of \monthyeardate\today. {\it Please cite published version, if available}}

\begin{document}

\maketitle

\setcounter{page}{1}

\begin{abstract}
    \singlespacing
    The value-alignment problem for artificial intelligence (AI) asks how we can ensure that the `values'---i.e., objective functions---of artificial systems are aligned with the values of humanity. In this paper, I argue that linguistic communication is a necessary condition for robust value alignment. I discuss the consequences that the truth of this claim would have for research programmes that attempt to ensure value alignment for AI systems---or, more loftily, designing robustly beneficial or ethical artificial {\it agents}.
    
    \phantom{a}
    
    \phantom{a}

\end{abstract}

\section{Introduction}
    \label{sec:Introduction}

    The value-alignment problem for artificial intelligence (AI) asks how we can ensure that the `values'---i.e., objective functions---of artificial systems are aligned with the values of humanity, writ large \citep{Russell-2019}. One component of this problem is {\it technical}, focusing on how to properly encode values or principles in artificial agents so that they reliably do what they ought to do---i.e., what {\it we} want them to do, or what we {\it intend} for them to do. Another component of value alignment is {\it normative}, emphasising what values or principles from normative theory are the `correct' ones to encode in AI systems \citep{Gabriel-2020}. %
    However, ensuring value alignment for AI systems---or, more loftily, designing robustly beneficial or `ethical' artificial {\it agents}---requires more than just translating our best normative theories into a programming language.

    I propose and defend the following claim, which I will refer to as `the {\bf main claim}' throughout this paper. 
            \begin{quote}
            \singlespacing
                {\bf Main Claim}.\\ Linguistic communication is a necessary condition for robust value alignment.
            \end{quote}
    I discuss the consequences of the {\bf main claim} in the specific context of value-alignment problems {\it for} artificial intelligence. However, one should note that the {\bf main claim} makes no specific reference to AI systems since I take it to hold in the context of value alignment more generally than just the specific instantiation seen in recent research on safe AI.

    In Section~\ref{sec:Concepts}, I flag some assumptions that I will make in this paper and clarify what is meant by each component of the {\bf main claim}, discussing linguistic communication (\ref{sec:Language}), value-alignment problems (\ref{sec:VAP}), and necessary conditions (\ref{sec:Necessity}) in turn. With this groundwork laid, I provide two arguments in favour of the {\bf main claim} in Section~\ref{sec:MainClaim}. The first (\ref{sec:Main1}) can be formulated explicitly as follows:
        \begin{enumerate}
            \small{\singlespacing
                \item Principal-agent problems are primarily problems of informational asymmetries.
                \item Value-alignment problems are structurally equivalent to principal-agent problems.
                \item {\bf Therefore}, value-alignment problems are primarily problems of informational asymmetries.
                \item Any problem that is primarily a problem of informational asymmetries requires information-transferring capacities to be solved, {\bf and} the more complex (robust) the informational burden, the more complex (robust) the information-transferring capacity is required.
                \item {\bf Therefore}, value-alignment problems require information-transferring capacities to be solved, {\bf and} sufficiently complex (robust) value-alignment problems require robust information-transferring capacities to be solved.
                \item Linguistic communication is a {\it uniquely} robust information-transferring capacity.
                \item {\bf Therefore}, linguistic communication is necessary for sufficiently complex (robust) value alignment.}
            \end{enumerate}
    The second argument (\ref{sec:Main2}) follows from the failures of the symbolic systems approach to AI in combination with the rigidity of objective functions for aligning values in present-day AI systems. Finally, I discuss some empirical evidence in favour of the {\bf main claim}~(\ref{sec:Main3}). In Section~\ref{sec:Discussion}, I highlight some logical implications of {\bf main claim}. Importantly, the {\bf main claim} specifies a lower bound on the difficulty of ensuring value alignment for AI. Section~\ref{sec:Conclusion} concludes.

\section{Conceptual Components of the Main Claim}
    \label{sec:Concepts}

    In this section, I provide some basic definitions and clarify some assumptions I will make in this paper. The {\bf main claim} contains three conceptual components---{\it linguistic communication}, ({\it robust}) {\it value alignment}, and the relation between them, {\it necessity}. These components are treated separately in Sections~\ref{sec:Language},~\ref{sec:VAP}, and~\ref{sec:Necessity}.

\subsection{Linguistic Communication}
    \label{sec:Language}

    Before defending the {\bf main claim}, the first thing to clarify is what is meant by `linguistic communication'. Linguistic communication, in this context, means something like `natural language'. That is to say, `linguistic communication' is understood as synonymous with `natural language', and I may use the two terms interchangeably throughout this paper.%
            \footnote{Of course, it is logically possible that there is a form of linguistic communication that is {\it not} a natural language, thus pulling these two concepts apart. However, natural language is a {\it prototypical} example of linguistic communication. As a result, in the context of this paper, we can understand them as relative synonyms.}
    Note that no {\it particular} linguistic communication system---e.g., English---will be necessary for value alignment in the context of the {\bf main claim}; instead, `linguistic communication' and `natural language' should be understood as a shorthand for the {\it abilities} that give rise to linguistic communication---i.e., the {\it capacity} for language. The capacities that give rise to linguistic communication systems can be contrasted with what we might call `simple communication' (`simple communicative abilities'), typified by the ubiquitous signalling systems in the natural world, from bacteria to bees to primates.%
            \footnote{See, e.g., \citet{Lishak-1984, Lugli-et-al-2003, Marler-Slabbekoorn-2004, Belanger-Corkum-2009, Houck-2009, Mathger-et-al-2009, Bruschini-et-al-2010, Costa-Leonardo-Haifig-2010, Haddock-et-al-2010, Wyatt-2010, Thiel-Breithaupt-2011}.}

    At the same time, it is perhaps too strong to require that all features of natural language are individually necessary for robust value alignment between agents. Rather, certain constitutive (and differentiating) features of linguistic communication will be necessary for robust value alignment. I will discuss more precisely which functions of linguistic communication are necessary preconditions below. So, not every feature of linguistic communication will be necessary for robust value alignment; however, those features that {\it are} necessary for robust value alignment are also the features of linguistic communication that make it (many researchers suppose) unique to humans.

    A few things are presupposed, which I will clarify but not defend at length. First, I assume that the primary {\it use} of natural language is {\it to communicate}. This view is distinct from the views of, e.g., \citet{Chomsky-1965, Chomsky-1980, Chomsky-1995}, who suggests that the primary purpose of language is the expression of thought.%
            \footnote{When Chomsky discusses `language' (or, more typically, `grammar'), he means `{\it I}-language'---i.e., a bio-computational system represented in the brain, with a capacity for generating a discrete infinity of hierarchical structures. See also, \citet{Bickerton-1990, Wray-1998, Hauser-et-al-2002}.}
    Those who think that language is not (primarily) for communication argue that it appears that only {\it some} aspects of language could be understood as adaptations {\it for} communication \citep[21]{Fitch-2010}. Thus, for Chomsky and those who follow the biolinguistic research programme---i.e., mainstream generative grammar---human language is identified as something like the FLN---the {\it faculty of language in the narrow sense} \citep{Hauser-et-al-2002}. The FLN emphasises the computational mechanisms required for (complex or compositional) syntax.%
            \footnote{\citet{Hauser-et-al-2002} hypothesise that the FLN is equivalent to recursion, which makes the FLN unique to {\it Homo sapiens}. This idea is sometimes referred to as the `recursion only' hypothesis---i.e., the hypothesis that recursion is {\it the} property that `distinguishes human language from animal communication systems' \citep[611]{Traxler-et-al-2012}. See criticism in \citet{Jackendoff-Pinker-2005, Pinker-Jackendoff-2005, Parker-2006, Wacewicz-2012}.}
    But, it has been argued that this `syntactocentrism'---where syntax is supposed to be the only generative component of language---is misplaced.%
            \footnote{See discussion in \citet{Jackendoff-1997, Jackendoff-2003, Jackendoff-2012, Culicover-Jackendoff-2005}. When the FLN is introduced by \citet{Hauser-et-al-2002}, its uniqueness to {\it Homo sapiens} is postulated as a hypothesis to be bolstered with empirical data. However, \citet{Fitch-et-al-2005} later {\it stipulate} that the FLN is unique to humans as a matter of definition. \citet{Wacewicz-2012} highlights that the description of the FLN proffered by \citet{Fitch-et-al-2005} is not merely a correction or clarification of the earlier definition given by \citet{Hauser-et-al-2002}; instead, they are conceptually inconsistent. Thus, the distinction between the {\it faculty of language in the broad sense} (FLB) and the FLN confuses, rather than clarifies, dialogue on language origins. \citet{Fitch-2017}, in particular, appears to have somewhat distanced himself from his early work with Hauser and Chomsky, referring instead to the {\it derived components of language} (DCLs), which include (at least) complex vocal learning, hierarchical syntax, and complex semantics/pragmatics. Thus, \citet{Fitch-2017} simultaneously de-emphasises the importance of syntax alone.}
    There is no obvious conceptual clarity to be gained by assuming that language is different {\it in kind} from communication. Furthermore, assuming that communication is the primary purpose of language is more parsimonious (in an evolutionary context) than alternative views, which assume it is not.%
            \footnote{See discussion in \citet{LaCroix-2020}.} 
    The view that language is primarily communicative further underscores the {\it continuity} between human linguistic abilities and the non-linguistic communicative abilities of non-human animals.

    This last point is relevant to discussions of value alignment {\it qua} moral behaviour insofar as an analogy holds between language and ethics. Briefly, linguistic ability and moral behaviour are typically considered unique to humans.%
            \footnote{For example, \citet{Kitcher-2011, Korsgaard-2018} argue that humans are the only animals with the sort of meta-cognitive capacities required for moral agency.}
    However, some non-human animals exhibit complex communication abilities, which contain features that might be called {\it proto-linguistic}.%
            \footnote{See discussion in \citet{Bickerton-1990, Bickerton-1995, Wray-1998, Wray-2000, Wray-2002, Jackendoff-1999, Jackendoff-2003, Arbib-2002, Arbib-2003, Arbib-2005, Tallerman-2007, Tallerman-2012, Planer-Sterelny-2021}.}
    Additionally, some non-human animals exhibit {\it proto-moral} behaviours in the form of the complex social structures, norms, and reactive emotions that give rise to robust cooperation (of the sort theorised by some as providing a foundation out of which human morality may have evolved).%
            \footnote{See discussion in \citet{Kitcher-2006a, Kitcher-2006b, Kitcher-2011, de-Boer-2011} and empirical examples given by \citet{de-Waal-1996, de-Waal-2006, Flack-de-Waal-2000}. \citet{Vincent-et-al-2019} argue that great apes participate in certain (foundational) normative practices like reciprocity, caring, social responsibility, and solidarity; similarly, \citet{Rowlands-2012} argues that some non-human animals can track moral truths. Of course, this debate may be semantic rather than substantive, as \citet{Fitzpatrick-2017} argues.}

    It is difficult to specify precisely what differentiates human-level linguistic communication (abilities) from simpler communication systems seen in non-human species. Nonetheless, certain distinctive features (or functions) of language, understood as a highly sophisticated communication system, will be necessary for the robust sort of value-alignment to which the {\bf main claim} refers. One such feature of linguistic communication is {\it compositionality}---and related features like hierarchy and recursion---which is often taken as a key differentiating component of linguistic communication systems insofar as it is absent in any hitherto studied animal communication system.%
            \footnote{See, e.g., discussion in \citet{Hauser-et-al-2002, Hauser-Fitch-2003, Mehler-et-al-2006, Fitch-2010, Hurford-2012, Scott-Phillips-Blythe-2013, Berwick-Chomsky-2016}. Of course, some non-human animals utilise remarkably complex communication systems---e.g., the combinatorial `waggle dance' of {\it Apis mellifera} \citep{Aristotle-History-IX, Spitzner-1788, von-Frisch-1967}; the syntactic songs of {\it Poecile atricapillus} \citep{Hailman-et-al-1985}; the `pyow-hack' signals of {\it Cercopithecus nictitans} \citep{Arnold-Zuberbuhler-2006b, Arnold-Zuberbuhler-2006a, Arnold-Zuberbuhler-2008, Arnold-Zuberbuhler-2013}; or the functionally-referential alarm calls of {\it Chlorocebus pygerythrus} \citep{Seyfarth-et-al-1980} and {\it Cercopithecus diana} \citep{Zuberbuhler-et-al-1999}, among others. Still, despite this remarkable variation and complexity, non-trivial (linguistic) compositionality appears to be unique to human-level linguistic communication systems.}
    `Compositionality', here, can be taken as the usual definition, which is given by the following principle:%
            \begin{quote}
                \singlespacing
                    {\bf Principle of (Linguistic) Compositionality}.\\
                    The meaning of a compound [complex] expression is a function of the meaning of its parts [constituents] and how they are combined [composed].%
                        \footnote{See \citet{Frege-1923, Partee-1984, Kamp-Partee-1995, Janssen-1997, Janssen-2012, Pagin-Westerstahl-2010-1, Pagin-Westerstahl-2010-2, Szabo-2012, Szabo-2020}. Of course, this formulation is inherently ambiguous: \citet{Szabo-2012} highlights that these ambiguities give rise to at least eight different readings of the principle.}
            \end{quote}
    The key thing here is that the elements of natural language can be combined into hierarchical phrases, which then may be recursively combined into larger phrasal expressions; and, the meaning of such an expression is a {\it function} of the meaning of its parts and how they are combined. The relation between recursion, hierarchy, and compositionality is such that recursion requires hierarchy---at least to some extent---and hierarchy requires compositionality---again, at least to some extent \citep{Beule-2008}. A key entailment of a communication system's being compositional is that, with a limited vocabulary and a finite set of grammatical rules, linguistic communication systems allow for the production and understanding of an unlimited number of unique expressions.

    The principle of (linguistic) compositionality contrasts with a notion of {\it trivial} compositionality, where one can always interpret the complex expressions of a communication system in terms of set intersection.%
            \footnote{See \citet{Schlenker-et-al-2016, Zuberbuhler-2018, Steinert-Threlkeld-2020}.}
    For example, the words `cute' and `dog' can compose to form the expression `cute dog' in English; however, the meaning of this expression is determined trivially by the intersection of the sets of {\it cute-things} and {\it dog-things}.%
            \footnote{See \href{https://travislacroix.github.io/atlas}{{\bf here}} for an exemplar.}
    The principle of (linguistic) compositionality also contrasts with the individual notions of {\it syntactic} and {\it semantic} compositionality. A complex signal is syntactically compositional just in case it is composed of atomic signals but is not interpreted compositionally. Conversely, it is semantically compositional just in case the meaning of the compositional signal can be {\it decomposed} in interpretation, even if the signal is holophrastic. Linguistic compositionality requires both semantic and syntactic composition.%
            \footnote{See discussion in \citet{LaCroix-2020, LaCroix-Information}.}

    Compositionality is typically understood as a requirement for explaining both the {\it systematicity} and {\it productivity} of linguistic communication.%
                \footnote{See discussion in \citet{Fodor-1998, Szabo-2020}.} 
        In any case, I will take it for granted that linguistic communication systems, typified by natural language, are compositional in a non-trivial way---i.e., the robust form of compositionality that appears to be a key feature of linguistic communication involves some {\it non-conjunctive}---or what we might call {\it functional}---modification of linguistic items by certain other linguistic items, such as function words \citep{Steinert-Threlkeld-2020}. Composition, hierarchy, and recursion are  sufficient for systematicity and productivity; so, compositionality, for the claims made in this paper, will be taken as a defining feature of linguistic communication.%
                \footnote{Note that this can be true regardless of whether the thing that evolved in {\it Homo sapiens} was compositionality per se. \citet{LaCroix-2020, LaCroix-2021, LaCroix-2019} argues that {\it reflexivity} is a more apt explanatory target for a (gradualist) evolutionary account of language origins. Reflexivity can give rise to compositionality as a byproduct.}

    To summarise, by `linguistic communication', I mean the particular {\it types} of communication systems that evolved in {\it Homo sapiens}. These species-unique communication systems involve (verbal or nonverbal) signals and their conventional meanings; they give rise to hierarchical, recursive, compositional structures; and, their primary function is to allow individuals to share information---i.e., to communicate. In short, linguistic communication systems are those communication systems that are typified by natural languages. I will discuss the importance of the non-trivially compositional features of linguistic communication systems for the {\bf main claim} in Section~\ref{sec:MainClaim}. First, I describe what is meant by `robust value alignment' in the {\bf main claim}.

\subsection{Robust Value Alignment} 
    \label{sec:VAP}

    The next thing to clarify is what is meant by `robust value alignment' in the context of the {\bf main claim}. I begin by discussing the {\it value-alignment problem} in its most general form, highlighting a range of conditions under which tokens of this type of problem may be generated (\ref{sec:GVAP}). I then turn to the value-alignment problem in the particular context of AI (\ref{sec:AIVAP}). Finally, I discuss the connection between the value-alignment problem and the {\it control} problem, arising in the context of superintelligent AI systems (\ref{sec:Control}). As in Section~\ref{sec:Language}, the main goal of this section is to clarify some basic concepts, definitions, and assumptions in the context of this paper. I will again postpone discussing the importance to the {\bf main claim} of the features of value-alignment problems discussed in this section to Section~\ref{sec:MainClaim}.

\subsubsection{General Value-Alignment Problems}
    \label{sec:GVAP}

    The value-alignment problem, in its most general form, is a problem of how two (or more) agents (actors) can align their values (or objectives). In economics, law, and politics, what I am calling the value-alignment problem is more commonly known as the {\it principal-agent problem}.%
            \footnote{This problem is sometimes referred to as an {\it agency dilemma} or an {\it incentive problem}; see discussion in \citet{Jensen-Meckling-1976, Eisenhardt-1989, Laffont-Martimort-2002}.} 
    This type of problem arises (or, at least, may arise) in any context where some entity---called `the principal'---appoints another entity---called `the agent'---to act on the principal's behalf. A problem may be generated, in part, because the principal and the agent have different objectives, incentives, or values. When there is a conflict of interest between the principal and the agent's values, we say their values are misaligned---hence this is a value-alignment problem. Thus, in its most general form, the {\it problem} of value alignment arises from the dynamics of multi-agent interactions involving the delegation of tasks from one actor to another.

    Some examples will clarify how ubiquitous this problem is in human-human interactions.
            \begin{enumerate}
            \singlespacing
                    \item A corporate executive, like a chief executive officer (CEO), runs a corporation on behalf of its shareholders. The shareholders are the principal(s), and the CEO is the agent.
                    \item[] 
                    \item In a democracy, citizens elect a representative (or representatives) who then act on behalf of the citizens. The citizens are the principal(s), and the elected party or officials are the agent(s).
                    \item[] 
                    \item A customer orders groceries online, and a grocer collects the items to be delivered. The customer is the principal, and the grocer is the agent. 
                    \item[] 
                    \item An individual's car performs sub-optimally, so they take it to a mechanic, who tells them several parts need to be replaced. The car owner is the principal, and the mechanic is the agent.
                    \item[] 
                    \item A homeowner uses a real estate agent to sell a house. The homeowner is the principal, and the real estate agent is the agent.
            \end{enumerate}%
    In case (1), we might imagine that the objective of the shareholders is for the corporation to maximise (shareholder) profit through increasing stock value. However, the CEO might use profits to proffer large bonuses to corporate-level executives instead of paying dividends to shareholders---an action that benefits the agent(s) but not the principal(s). In case (2), citizens might vote for a party whose platform (appears to) align with their values; however, once elected, the party may renege those promises in light of competing considerations that benefit the party rather than the citizens. In case (3), we might imagine that a customer values certain properties in their groceries---e.g., freshness---whereas the grocer values certain other things---e.g., getting rid of items close to expiration. In case (4), the car-owner may value having all and only the required work performed, and the mechanic may value maximising unnecessary expense to their own benefit. In case (5), the homeowner may value getting the best price on the house, whereas the agent may value closing the sale as quickly as possible (even at a reduced price); or, the agent may over-promise on the price to win the listing in the first place, simultaneously making it more difficult to sell the property.

    A principal-agent problem can arise in each case because the principal and the agent may have different values, interests, or objectives. However, another key feature common to these examples is {\it informational asymmetry}.%
            \footnote{In the 1970s, economists showed how asymmetric information poses significant challenges to {\it General Equilibrium Theory} \citep{Akerlof-1970, Spence-1974, Rothschild-Stiglitz-1976}. See also \citet{Marschak-Radner-1972} and discussion in \citet{Laffont-Martimort-2002}.} 
    In case (1), the shareholders do not have information about the day-to-day goings-on of the business, but the CEO does. In case (2), the citizens use platforms as a proxy for choosing the candidate whose values most align with their own while not knowing which values the party will instantiate once elected. In case (3), the principal cannot observe the agent's actions. In cases (4) and (5), the principal lacks specialised information, which allows the agent to take advantage for their own gain. Informational asymmetries and imperfect information contribute to the generation of principal-agent problems.%
            \footnote{Note that `perfect information' is a term of art. In Economics, perfect information implies that all market participants have all the information required to make a decision. In game theory, perfect information means that a player knows the game's entire history up to the decision point, as in backgammon. Imperfect information is the negation of perfect information; this occurs when some information is unavailable or hidden. Thus, imperfect or incomplete information means that there is some uncertainty. See discussion in \citet{Neumann-Morgenstern-1944, Shapley-1953}.}

    \citet{Laffont-Martimort-2002} highlight that if there is no private information between a principal and an agent, then {\it even if} the agent's objectives conflict with the principal's, the principal could still propose `a contract which perfectly controls the agent and induces the [agent's] actions to be what he [the principal] would like to do himself in a world without delegation' (12). Essentially, under complete information, the principal has complete {\it bargaining power}.%
        \footnote{In classical game theory, this is often operationalised as a higher disagreement point for the powerful agent; see discussion in \citet{LaCroix-Oconnor-2021}.}
    Therefore, misaligned values alone are insufficient for generating a principal-agent problem since they can be controlled when there is no informational asymmetry between the principal and the agent.%
        \footnote{This claim is mathematically provable on the particular economic model we are discussing. Of course, since this is a model and, therefore, an idealisation, we might question whether this model is sufficiently applicable to real-world interactions and whether this claim holds in the real world.}
    There are three different ways that private information can generate an agency dilemma.

    First, {\it hidden knowledge} is an informational asymmetry resulting from the agent's private information---e.g., concerning their own skills or opportunity costs---to which the principal does not have access. Hidden knowledge may contribute to the generation of a principal-agent problem.%
            \footnote{In economics, this is referred to as {\it adverse selection}. Hidden knowledge on the agent's part causes the principal to give up some information rent. Thus, a contract must be designed to elicit private information, which may be costly to the principal. See \citet{Akerlof-1970, Rothschild-Stiglitz-1976, Spence-1973, Spence-1974, Laffont-Martimort-2002, Hou-et-al-2009}.}
    For example, a prospective employee (the agent) knows their background skill level and appropriateness for a particular job, whereas the hiring committee (the principal) does not. In cases of hidden knowledge, uncertainty is exogenous to the relationship between the principal and the agent. Cases (4) and (5) above are examples of informational asymmetries arising from hidden knowledge.

    Second, {\it hidden action} is an informational asymmetry caused by the agent's ability to perform an action that the principal cannot observe.%
            \footnote{In economics, this is known as {\it moral hazard}. See \citet{Haynes-1895, Knight-1921, Arrow-1963, Arrow-1968, Vaughan-1997, Laffont-Martimort-2002}.}
    When the risk-taking individual (the agent) knows more about their intentions than the consequence-paying individual (the principal), the agent may take on more risk than the principal would otherwise be comfortable with, as is common with insurance. In this case, the uncertainty due to asymmetric information is endogenous to the relationship of the principal and the agent. Cases (1) to (3) above exemplify informational asymmetries arising from covert action.

    Solutions to principal-agent problems arising from informational asymmetries due to hidden action and hidden knowledge assume that information is (ex post) verifiable by an independent third party, such as a (benevolent) Court of Justice \citep{Laffont-2000}. However, a third type of informational asymmetry may give rise to an agency dilemma when we assume that the information between an agent and principal is symmetric (ex post) but unverifiable {\it in principle} by a third party \citep{Sappington-1991}.  Thus, the third type of informational asymmetry that may generate a principal-agent problem arises from the non-verifiability of (otherwise symmetric) information.%
            \footnote{Non-verifiability is particularly relevant in the field of {\it contract law}, though \citet{Shah-2014} notes that non-verifiability receives much less coverage in the economic literature on principal-agent problems than hidden knowledge (adverse selection) and hidden action (moral hazard). See \citet{Williamson-1973, Williamson-1975, Grossman-Hart-1986, Sappington-1991, Hart-1995, Laffont-2000, Laffont-Martimort-2002}.}
    For example, we might suppose that the principal and the agent have identical and sufficient (ex ante) information to complete some transaction. Here, a principal-agent problem may still arise when the agent represents the true state of the world as being different than they (and the principal) know it to be. This is a problem when either the agent's representation of the world is unverifiable by a third party {\it or} when it is too costly for a third party to verify.

    Thus, value misalignment may give rise to a principal-agent problem. However, when information is symmetric, the principal can create a contract that induces the agent to act just as the principal would in the absence of delegation. Therefore, value misalignment {\it alone} is not sufficient to generate a principal-agent problem. 

    Furthermore, I suggest that value misalignment is also {\it unnecessary} to generate a principal-agent problem. %
    For example, suppose the agent and principal have perfectly-aligned objectives, but they have no way to transmit that information. In that case, it is still possible that the agent's actions misalign with the principal's objectives (despite the agent's intentions). This situation might occur if, for example, there is an optimal action which would satisfy the principle's objectives---and, {\it ex hypothesi}, would also satisfy the agent's objectives---but the existence of this action is not common knowledge.
            \footnote{Suppose the philosophical literature on peer disagreement is to be trusted. In that case, it is at least possible for this to be true even when both parties are privy to identical evidence.} 
    Simple coordination games, where the agent chooses an action and the principal and agent both receive the same payoff, provide a clear example of this possibility.%
            \footnote{As a toy case, suppose the principal chooses a number, and the agent has to guess the correct number for both players to receive payoff. Their values are perfectly aligned, but the agent may still act in a way that the principal would not, precisely because there is an informational asymmetry between them. (With thanks to Aydin Mohseni for raising this possibility to me.) This toy example is much less artificial than it may seem at first glance---simple signalling games have a similar structure.}

    To summarise, a principal-agent {\it problem} may arise in any situation where an entity (the agent) can make decisions or take actions on behalf of, or that impact, another entity (the principal). The delegation of tasks from the principal to the agent generates a problem of managing information flows. Therefore, principal-agent problems are fundamentally problems of asymmetric information rather than asymmetric values (though value misalignment may exacerbate these problems).%
        \footnote{The preceding discussion should make clear that I am using certain terms as relative synonyms---for example, `objectives', `values', and `interests'. Although there may be relevant conceptual distinctions between these terms, I will ignore these for ease of exposition in this paper. When I discuss artificial agents, I will refer to `their values' without worrying too much about anthropomorphism; I will take the `values' of an AI system to mean something like the `objectives'---in the sense of {\it objective functions}---of said system.}

\subsubsection{Value Alignment Problems for AI}
    \label{sec:AIVAP}

    In the preceding discussion of principal-agent problems from economics, each example involved interactions between a human principal and a human agent. However, this class of problems also arises in the context of artificial agents or AI systems, where it is typically referred to as the `value-alignment problem'. \citet[417]{Hadfield-Menell-Hadfield-2019} suggest that the value-alignment problem has a `clear analogue' in  principal-agent problems. However, we can make a stronger claim: the relationship between value-alignment problems for AI and principal-agent problems in an economic context is not an {\it analogy} but {\it identity}---at least in terms of formal structure. However, the structural identity between these problems does not imply that their solutions will be identical, as we will see.%

    For clarity, in the remainder of this paper, I will use `the principal-agent problem' to refer to the class of problems arising from conflicts of interest and asymmetric information in general interactions between two entities, as discussed in Section~\ref{sec:GVAP}. I will use `the value-alignment problem' to refer to the subset of principal-agent problems arising in human-AI interactions.

    `Artificial intelligence' refers to a property of an artificial system–––i.e., that it `thinks', `acts', or `behaves' in an intelligent way. Intelligence, in this context, is best understood as `an agent's ability to achieve goals in a wide range of environments' \citep[12]{Legg-Hutter-2007}.%
            \footnote{See also \citet{Gardner-2011, Cave-2017, Gabriel-2020, Russell-Norvig-2021}.}
    `AI' also refers to an approach or set of techniques for {\it achieving} this property in an artificial system \citep{Gabriel-2020}. The most promising method for achieving machine intelligence in recent years is machine learning (ML). This approach to AI involves training models using (typically huge amounts of) data. The models `learn' gradually to behave in the desired way without that behaviour being explicitly programmed. The three main techniques for ML today are supervised learning, unsupervised learning, and reinforcement learning \citep{Goodfellow-et-al-2016}.

    In supervised learning, a model is trained on labelled examples---the {\it training data}. The model is evaluated on how well it generalises what it learns from the training data to previously unseen examples---the {\it test data}. In unsupervised learning, an algorithm learns underlying patterns or correlations from unlabelled data. Reinforcement learning (RL) depends upon sparse rewards for actions \citep{Sutton-Barto-2018}. Since 2012, the main driver of AI research has been {\it deep learning} (DL). This approach to ML utilises deep neural networks modelled (roughly) after neurons in the human brain \citep{Savage-2019}. DL uses layers of algorithms to process data---information is passed through each subsequent layer in a neural network, with the previous layer's output providing input for the following layer. One of the key advantages of DL techniques is that they do not require the heavily hand-crafted features used by traditional methods for AI \citep{Buckner-2019}.

    ML, at its core, is just an {\it optimisation problem}. The thing being optimised, in this case, is an {\it objective function}, which gives a way to configure the system to bring it closer to the {\it ground truth} provided during model training. Essentially, an objective function provides a {\it proxy} for {\it what we want the system to do}. Consider an example from supervised learning for image recognition. Given a dataset, ${\bf D}$, of image-label pairs, $(x,y)$, the {\it true objective}---i.e., what we want or intend for the system to do---is to have the model correctly guess the image label for previously unseen images. The objective {\it function}, in this case, might be a probability distribution function, %
        $$p(\hat y = y\,|\, x, \theta),$$
    which outputs the conditional probability that the predicted label, $\hat y$, is identical to the true label, $y$, given the observed data, $x$, and a model, $\theta$.
    
    The true label, $y$, determines whether a particular predicted label, $\hat y$, is correct for a particular image, $x$. A {\it loss function} is used to optimise the model, $\theta$. According to a specified evaluation metric---e.g., mean-squared error%
            \footnote{The mean squared error is given by $\frac{1}{n} \sum_{i=1}^{n}(Y_{i} - \hat{Y}_{i})^{2}$, where $n$ is the number of data points, $Y_{i}$ is the set of observed values, and $\hat{Y}_{i}$ is the set of predicted values.}%
    ---the loss function tells us how close the model is to the correct prediction for a single point.%
        \footnote{The loss function is sometimes differentiated from a {\it cost function}, which tells us the {\it average} loss over the entire dataset. This is sometimes couched in the language of {\it empirical risk}, which describes the loss over all samples in the dataset, which is then contrasted with the {\it true risk}---i.e., the loss over all samples. More accurately, an objective function approximates our true objective, and the objective function determines the empirical risk, which approximates the true risk. See discussion in \citet[Sec. 8.1]{Goodfellow-et-al-2016}.}
    Thus the {\it true objective}---correctly labelling images---is approximated by an {\it objective function}, which determines a metric for how close the model is to the objective.%
        \footnote{In fact, this {\it approximates} an approximation. (With thanks to Michael Noukhovitch for bringing this to my attention). Loss is measured on the test set, which gives a measure of generalisation error for a given objective function. Hence, optimisation actually occurs on the training set. Still, the hope is that we are also optimising the test set---i.e., achieving some degree of generalisation. See discussion in \citet[Sec. 5.2]{Goodfellow-et-al-2016}.}
    If loss is close to zero, then the model successfully optimises the objective function according to the metric used. Further, if the objective function is a good proxy for the true objective, then optimising an objective function means optimising the objective. Therefore, a given system's objective function(s) must be accurately specified \citep{Reed-Marks-1999}.

    In the context of AI, the value-alignment problem can arise when objectives are misspecified. Part of the difficulty here arises because objectives require programmers to define an objective function, which can be difficult to operationalise in a programming language. Additional complications arise from the fact that objective {\it functions} are mere proxies for the {\it true} objective; however, the objective function is identical to the true objective from the AI system's `point of view'. When objective functions are poorly specified, this will lead to a value-alignment problem insofar as there is a misalignment between the actual objective---our `values'---and its proxy---the system's `values'.

    The provision of an objective function implicitly defines an optimisation `landscape' \citep{Sipper-et-al-2018}; however, the solution space for an optimisation problem defined by an objective function  includes a `pathology' of {\it local} optima \citep{Lehman-Stanley-2008}, meaning that these landscapes are frequently `deceptive' \citep{Goldberg-1987, Mitchell-et-al-1992}. \citet{Lehman-Stanley-2008} highlight that the objective function `does not necessarily reward the stepping stones in the search space that ultimately lead to the objective' (329), meaning that objective functions are often constructed {\it ad hoc}. Thus, poorly-designed objective functions can lead to value-alignment problems whenever there is a conflict between the objective function and the actual objective. Furthermore, even when objective functions are accurate proxies, sufficiently complex action spaces will have local optima that may result in outputs that are grossly misaligned with the true objective. A mismatch between a well-specified objective function and the emergent behaviour of an AI system is sometimes referred to as `inner alignment'; when the objective is misspecified, this is referred to as `outer alignment' \citep{Hubinger-et-al-2021}.%
        \footnote{See also discussion in \citet{Ecoffet-et-al-2020, Christian-2020, Krakovna-et-al-2021, Clark-Amodei-2016}.}

    These value-alignment problems fall under the heading of `AI safety'. \citet{Amodei-et-al-2016} highlight that when objective functions are misspecified, this may give rise to unanticipated side effects or reward hacking; when objective functions are too expensive to evaluate at regular intervals, this creates a problem of scalable supervision; and, local optima of objective functions may lead to undesirable behaviour during learning.%
        \footnote{\citet{Amodei-et-al-2016} refer to these problems as `accidents', but how they can be classed as value-alignment problems should be clear. See discussion in \citet{Hadfield-Menell-Hadfield-2019}.}
    These problems are exacerbated because the utilities determined by any concretely-specified objective function will necessarily be a mere {\it subset} of our utilities---i.e., the things we value \citep{Yudkowsky-2018}. This difficulty leads to (what \citet[p. 9]{Russell-2019} sees as) the failure of the `standard model' of intelligence for AI systems. Namely, a standard definition of intelligence in humans might be formulated as follows: %
        \begin{quote}
        \singlespacing
            {\it Humans are \textbf{intelligent} to the extent that \textbf{our} actions can be expected to achieve \textbf{our} objectives}. 
        \end{quote}
    Given this definition, the `standard model' for {\it machine} intelligence has been given analogously, thus: %
        \begin{quote}
        \singlespacing
            {\it Machines are \textbf{intelligent} to the extent that \textbf{their} actions can be expected to achieve \textbf{their} objectives}.
        \end{quote}
    The problem is that, unlike humans, machines have no objectives of their own. Thus, we must define their objectives, which gives rise to the possibility of value misalignment.

    Each of the problems described above involves encoding objectives in an AI system. \citet{Gabriel-2020} refers to this as the `technical component' of the value-alignment problem---namely, how do we encode values, principles, objectives, etc. in AI systems so that they `reliably do what they ought to do' (412), or what we {\it intend} for them to do. \citet{Gabriel-2020} further distinguishes the technical component of the value-alignment problem from the `normative component', which involves the problem of determining {\it what} values (objectives) should be encoded in an AI system in the first place.%
            \footnote{It is worth noting that \citet{Gabriel-2020} uses the language of `artificial agent' rather than `AI system' when discussing value alignment. A `standard' philosophical account of agency requires something like {\it intentional action}. This view is defended by, e.g., \citet{Davidson-1963, Davidson-1971, Goldman-1970, Brand-1984, Bratman-1987, Dretske-1988, Bishop-1989, Mele-1992, Mele-2003, Enc-2003}, among others. See discussion in \citet{Schlosser-2019}. 
        However, even a more inclusive notion of agency is unnecessary for generating value-alignment problems. As these systems are further integrated into society, this problem becomes more pressing \citep{LaCroix-Bengio-2019}.}
    
    Part of the idea is that as these systems become more integrated into society, some of their decisions may carry moral weight so that we might classify their actions as `moral' or `immoral'. These considerations have given rise to the field of {\it machine ethics}, which seeks to `implement moral decision-making faculties in computers and robots' \citep{Allen-et-al-2006}.%
        \footnote{See \citet{Tolmeijer-et-al-2020} 
        for a recent survey of machine ethics and approaches to `artificial moral agency' (AMA).}

        Value-alignment problems are truly ubiquitous. `Accidents', like those described in \citet{Amodei-et-al-2016}, are (technical, internal) value-alignment problems to the extent that misspecified or costly objective functions may lead to behaviour misaligned with what we intended the system to do. More generally, machine bias and problems arising from fairness considerations are value-alignment problems---at least to the extent that we do not want or intend for models to act in ways that we would call `racist', `sexist', or otherwise discriminatory.%
            \footnote{Some salient examples are described in, e.g., \citet{Angwin-et-al-2016, Christian-2020, Tomasev-et-al-2021, Miceli-et-al-2022}.}

        Note that all the problems mentioned have the same structure as the more general {\it principal-agent problems} described in Section~\ref{sec:GVAP}. Although misaligned values (objectives) can exacerbate these problems, the real cause of a value-alignment problem is asymmetric information. Complex computational systems, like those that undergird AI systems, are often `ineliminably opaque' \citep[568]{Creel-2020}. Opacity implies that verifying alignment for sufficiently complex systems or situations may be impossible. This difficulty is worsened when there is no simple matter of fact about what the system should align with in the first place.%
            \footnote{See discussion in \citet{LaCroix-Forthcoming, LaCroix-Luccioni-2022}.} 
        These considerations echo the insight from economics that `by definition the agent has been selected for his specialized knowledge and the principal can never hope to completely check the agent’s performance' \citep[538]{Arrow-1968}. Therefore, {\it transparency}, {\it interpretability}, and {\it explainability} \citep{Creel-2020, Creel-2021, Erasmus-et-al-2021, Kasirzadeh-2021} are also fundamentally problems of value alignment. These considerations further underscore the insights discussed in Section~\ref{sec:GVAP} that {\it informational asymmetries} generate principal-agent problems.

        Essentially, {\it any} problem endogenous to the model can be considered a value-alignment problem.%
                \footnote{Considering the wider class of interactions that arise from principal-agent problems, negative consequences exogenous to the model might also be called value-alignment problems. However, these will arise primarily from interactions between users of these systems and the stakeholders affected. The prototypical example is {\it misuse}, where a bad-faith actor intentionally uses technology for harm. However, endogenous value misalignment can be more subtle---as when a research group incentivised by, e.g., profit creates a system that adversely affects some subset of society. See discussion in \citet{Falbo-LaCroix-2021} for some examples.}
        Even problems of {\it control}, which arise primarily in the context of superintelligent AI, are a type of value-alignment problem insofar as a lack of control of a system that is capable of modifying its own reward function implies a misalignment of values (again, to the extent that we value maintaining control over such a system).%
                \footnote{See discussion in \citet{Bostrom-2014, Soares-et-al-2015, Orseau-Armstrong-2016, Hadfield-Menell-Hadfield-2019, Russell-2019}.}
        As it turns out, the problem of value alignment and the problem of control are closely connected.

\subsubsection{The Connection Between Value Alignment and Control}
    \label{sec:Control}

    \citet{Bostrom-2014} asks the following question: `how can the sponsor of a project that aims to develop superintelligence ensure that the project, if successful, produces a superintelligence that would realize the sponsor's goals?' (55). This is clearly a question of value alignment, where the principal is a project sponsor, and the agent is a superintelligence. \citet{Bostrom-2014} acknowledges the connection between this particular value-alignment problem and principal-agent problems from economics. The problem can be divided into two components. The first, which \citet{Bostrom-2014} calls `the first principal-agent problem', is a principal-agent problem centred on human-human interactions, where the principal is the sponsor of a research project aimed at creating a superintelligence and the agent is the developer of that project. \citet{Bostrom-2014} notes that the sponsor (the principal) could be a single individual or the entirety of humanity. The important thing is that the sponsor delegates authority to the developer, giving rise to a principal-agent problem.

    However, a different problem arises from the interaction between the {\it project} and the {\it system}---i.e., the superintelligence the project creates. This is the {\it control problem}, which \citet{Bostrom-2014} also refers to as `the second principal-agent problem'. Effectively, the principal must ensure that the intelligent system it creates will not act in ways that harm the principal's interests. \citet{Bostrom-2014} highlights that `standard management techniques'---i.e., those discussed in Section~\ref{sec:GVAP}---apply to the first principal-agent problem. In contrast, managing the second principal-agent problem will require brand-new techniques.%
            \footnote{Some such techniques are discussed in Chapter 9 of \citet{Bostrom-2014}.}

    Whereas the first-principal agent problem is ubiquitous, the second principal-agent problem arises only in the context of a {\it superintelligent} AI system. Control is a problem because several instrumental goals---e.g., self-preservation, goal-content integrity, cognitive enhancement, resource acquisition, etc.---are useful sub-goals for almost any original objective we may have programmed in the system \citep[141]{Russell-2019}.%
        \footnote{\citet{Bostrom-2014} refers to this as the {\it instrumental convergence thesis}:%
            \begin{quote}
            \singlespacing
                {\bf The Instrumental Convergence Thesis} \\
                Several instrumental values can be identified which are convergent in the sense that their attainment would increase the chances of the agent's goal being realized for a wide range of final goals and a wide range of situations, implying that these instrumental values are likely to be pursued by a broad spectrum of situated intelligent agents. (132)
            \end{quote}}
    Thus, one need not explicitly program incentives for self-preservation in an AI system.

    However, the control and value-alignment problems are closely related, as \citet{Wiener-1960} highlights:
        \begin{quote}
        \singlespacing
            If we use, to achieve our purposes, a mechanical agency with whose operation we cannot efficiently interfere once we have started it, because the action is so fast and irrevocable that we have not the data to intervene before the action is complete [i.e., the control problem], then we had better be quite sure that the purpose put into the machine is the purpose which we really desire and not merely a colorful imitation of it [i.e., the value-alignment problem]. (1358)
        \end{quote}
    Essentially, if we cannot solve the control problem before the advent of a sufficiently powerful AI system, we had better have solved the value-alignment problem before this happens so that the system's uncontrolled (or uncontrollable) actions at least align with our values.%
        \footnote{Of course, value drift or the possibility of such an artificial system being able to change its values implies that solving the value-alignment problem alone will not suffice to ensure safe and beneficial AI.}

    Thus, control problems are a type of value-alignment problem. The logical relation between these is that value alignment is necessarily prior to control: value alignment is a problem even for `narrow' or `weak' AI systems, whereas control is a problem only for more robust AI systems---e.g., artificial general intelligence or `strong' AI.

\phantom{a}

    To summarise, principal-agent problems arise whenever (1) an entity acts on behalf of another and (2a) their values are misaligned or (2b) there are informational asymmetries between them. In an economic context, the actors are typically described as individual (human) agents, government bodies, corporations, etc. However, this is not a logical requirement. The agent may be artificial, in which case the problem is typically referred to as a `value-alignment problem'. The value-alignment problem is the problem of ensuring that an AI system is properly aligned with human values. This problem is functionally equivalent to the (more general) principal-agent problem from economics, which can arise in any situation where a principal delegates authority to an agent who acts on the principal's behalf. This specification of a principal-agent problem is the situation we find ourselves in with the increasing autonomy of AI systems today. Importantly, the functional equivalence of principal-agent and value-alignment problems implies that informational asymmetries are the main driver of the latter since value misalignment alone is neither necessary nor sufficient for the former. Finally, value-alignment problems are ubiquitous to AI systems, regardless of how narrow they are. Therefore, problems of value alignment are necessarily prior to problems of control, which arises only in the context of a sufficiently strong AI system.

\subsection{Necessary Conditions}
    \label{sec:Necessity}

    Before defending the {\bf main claim} in Section~\ref{sec:MainClaim}, the final thing to clarify is the logical connection between linguistic communication and robust value alignment---i.e., {\it necessity}. Namely, the {\bf main claim} says that linguistic communication (as described in Section~\ref{sec:Language}) is {\it necessary} for robust value alignment between actors (as described in Section~\ref{sec:VAP}). `Necessary' is meant in the bread-and-butter sense of the material conditional.%
            \footnote{That is, $P \rightarrow Q$.}
    Contrapositively, one cannot achieve the robust form of value alignment required for safe and beneficial artificial intelligence without linguistic communication.

    Thus, three logically equivalent ways of specifying the {\bf main claim} quasi-formally in English are as follows:
            \begin{quote}
            \singlespacing
                (1) {\bf Main Claim}.\\ Linguistic communication is {\it necessary} for (the possibility of) robust value alignment.
            \end{quote}
            \begin{quote}
            \singlespacing
                (2) {\bf Alternative Claim}.\\ Robust value alignment (between actors) is possible {\it only if} those actors can communicate linguistically.
            \end{quote}
            \begin{quote}
            \singlespacing
               (3) {\bf Contrapositive Claim}.\\ Without natural language, robust value alignment is not possible. 
            \end{quote}
    The equivalent formulation of the {\bf main claim} that (2) proffers helps to clarify what is meant by `robust'. Namely, we can understand value alignment as a matter of {\it degrees}---from perfectly misaligned to perfectly aligned. Thus, (2) underscores that {\it some} communicative ability (in the sense of information transfer discussed in Section~\ref{sec:Language}) will be necessary for {\it some degree} of value alignment. This is discussed further in Section~\ref{sec:Main1}.

    Therefore, `robust' value alignment can be understood to specify a situation in which two actors' values (objectives) are sufficiently close to perfectly aligned. How close `sufficiently close' needs to be for `robust' value alignment will depend upon the context to which the values are relevant. Effectively, when the stakes are arbitrarily high, values will need to be arbitrarily close to perfectly aligned to say that the alignment of values is robust. In situations where the stakes are relatively low, it may be tolerable for values to be somewhat misaligned (again, relative to the context in question), but this would still be sufficient to claim robustness. Thus, the `robust' component of robust value alignment is intentionally vague since the requirement will be highly sensitive to context.

    However, part of the impetus for discussing value-alignment problems in the context of artificial intelligence is that the stakes appear to increase as these systems become more powerful---hence the importance of the {\bf main claim}, defended in the subsequent section.

\section{The Main Claim, Defended}
    \label{sec:MainClaim}

    Given the clarifications from Section~\ref{sec:Concepts} about what is meant by `linguistic communication', `robust value alignment', and the necessary connection between the two, we can now see why the {\bf main claim} might be plausible. I provide two arguments in favour of the {\bf main claim}. The first argument hinges on the unique advantages of linguistic communication for information transfer (Section~\ref{sec:Main1}). The second argument proceeds from the inherent rigidity of present-day objective functions and their formulation (Section~\ref{sec:Main2}). Neither of these arguments is definitive; however, I take them to underscore the plausibility of the {\bf main claim}. That said, in Section~\ref{sec:Main3}, I provide additional empirical evidence in favour of the {\bf main claim}. Section~\ref{sec:Discussion} examines some consequences if the {\bf main claim} is true.

\subsection{Language, Value Alignment, and Information Transfer}
    \label{sec:Main1}

    The first argument in favour of the {\bf main claim} follows from the joint realisation that (1) value-alignment problems are structurally equivalent to principal-agent problems, discussed in Section~\ref{sec:AIVAP}; (2) principal-agent problems are fundamentally problems of information transfer, discussed in Section~\ref{sec:GVAP}; and (3) linguistic communication, owing to features of systematicity and generalisability, is a uniquely robust and flexible communication system which allows for the transfer of information to an arbitrary degree of specificity, as discussed in Section~\ref{sec:Language}.

    Thus, the simplest way of understanding why we should expect the {\bf main claim} to hold is as follows:
        \begin{enumerate}
        \small{\singlespacing
            \item Principal-agent problems are primarily problems of informational asymmetries.
            \item Value-alignment problems are structurally equivalent to principal-agent problems.
            \item {\bf Therefore}, value-alignment problems are primarily problems of informational asymmetries.
            \item Any problem that is primarily a problem of informational asymmetries requires information-transferring capacities to be solved, {\bf and} the more complex (robust) the informational burden, the more complex (robust) the information-transferring capacity is required.
            \item {\bf Therefore}, value-alignment problems require information-transferring capacities to be solved, {\bf and} sufficiently complex (robust) value-alignment problems require robust information-transferring capacities to be solved.
            \item Linguistic communication is a {\it uniquely} robust information-transferring capacity.
            \item {\bf Therefore}, linguistic communication is necessary for sufficiently complex (robust) value alignment.}
        \end{enumerate}
    Premise (1) was argued for in Section~\ref{sec:GVAP}. Essentially, there are four possible combinations of values and information between a single principal and a single agent: either their values (objectives) are aligned, or they are misaligned, and either there are informational asymmetries between the actors, or there are not. So, we can ask which of the four combinations can possibly give rise to an agency dilemma. 
    
    Suppose the objectives of the principal and the agent are perfectly aligned, and there is no hidden information. In that case, no principal-agent problem is generated---the agent has the same values and information as the principal, meaning that if the agent acts in her own best interest, she will instantiate the principal's objectives by default. If objectives are misaligned, and there is perfect information, then there is {\it still} no problem generated, as we have seen.%
        \footnote{Recall that if objectives are misaligned, but there is no private information, then `the principal could propose a contract which perfectly controls the agent and induces the latter's actions to be what he would like to do himself in a world without delegation' \citep{Laffont-Martimort-2002}.} 
    It is only {\it possible} to generate an agency dilemma with imperfect information. As argued in Section~\ref{sec:GVAP}, this can happen even when objectives are perfectly aligned. So, the information component of the interaction generates the principal-agent problem, not the alignment or misalignment of objectives. The four possible situations are summarised in Table~\ref{tab:Combinations}. %
        \begin{table}[htb!]
        \centering
            \begin{tabular}{cc|cc}
                \multicolumn{2}{c}{} & \multicolumn{2}{c}{{\bf Values}} \\
                \multicolumn{4}{c}{} \\
                & & Aligned & Misaligned \\
                \cline{2-4}
                \multirow{2}{*}{{\bf Information} $\quad$} & Perfect & No & No \\
                & Imperfect & Yes & Yes \\
                \multicolumn{4}{c}{} \\
            \end{tabular}
        \caption{Combinations of information and values, giving rise to possible agency dilemmas (or not)}
        \label{tab:Combinations}
        \end{table}

    The key thing to note is that {\it informational asymmetries} give rise to principal-agent problems, at least in an economic model. 
    One may worry that this view conflates all incompleteness problems to information. The problem of misaligned objectives is dissolved when it is possible to write a complete contract, and informational asymmetries are fundamental in the real-world inability to specify complete contracts. However, it is also essential that arbitrary contracts include adequate {\it incentives} or {\it penalties} to properly shift the agent's objectives to align with the principal's objectives.%
        \footnote{Thanks to Gillian K. Hadfield for raising this point to me.} 
    The question of shifting objectives is raised in Section~\ref{sec:Main2}; however, the important thing to note is that {\it understanding} the incentives or punishments is a prerequisite for incentives or punishments to shift objectives. Whether this understanding requires robust information transfer, in general, is questionable---rewards and punishments in basic reinforcement learning may align {\it simple} objectives via, e.g., food or a shock. However, as the environment and the values that agents carry in those environments become more complex, the difficulty of {\it specifying}---let alone aligning---these values increases exponentially. This can be seen in many real-world examples from AI research. Although reinforcement learning can be very useful for complex environments where objectives are difficult to define, it is also difficult to specify rewards; \citet{Sumers-et-al-2022} highlight that linguistic communication `is an intuitive and expressive way to communicate reward information to autonomous agents' (1). And, as was discussed in Section~\ref{sec:AIVAP}, step-wise progress toward an ultimate goal may lead to unexpected or undesirable behaviour.

    Premise (2) was argued for in Section~\ref{sec:AIVAP}. As we have seen, the value-alignment problem for artificial intelligence is structurally equivalent to the principal-agent problem insofar as the {\it form} of the value-alignment problem arises from the dynamics of multi-agent interactions involving the delegation of tasks from one actor to another. `The value-alignment problem', in the context of AI systems, is just a type of principal-agent problem where the agent is an artificial system. The sub-conclusion given in Premise (3) follows logically from Premises (1) and (2).%

    The first conjunct of Premise (4) should be relatively uncontroversial. If a problem fundamentally involves informational asymmetries, it requires {\it some} ability to transfer information for its solution. The second conjunct of Premise (4) suggests that the necessary resources for information transfer effectively scale with the complexity of the problem. As the problem of aligning values becomes more complex, coordination on those values requires more sophisticated information-transferring capacities on the part of the actors involved in the coordination problem.%
            \footnote{Using the signalling-game framework, \citet{Steinert-Threlkeld-2016} argues that compositional signalling is only evolutionarily beneficial when the world is sufficiently complex. It should be unsurprising that the structural properties of language are affected in non-trivial ways by the world in which those language evolve \citep{Barrett-LaCroix-2022}.}

    In the language-origins literature, the {\it co}-evolution of coordination (cooperation) and communication explains why language {\it only} evolved in the hominin lineage \citep{Sterelny-2012, Planer-Sterelny-2021}. Considering the informational burden of robust cooperative capacities, \citet{Planer-Sterelny-2021} highlight that 
            \begin{quote}
            \singlespacing
                the shift [in the hominin lineage] to delayed-return cooperation introduced a range of new social challenges. It is at this stage that tracking the reputation of third parties via gossip or its equivalent became essential to the stability of cooperation, which drove the need for more complex communicative technologies. (xiv)%
                    \footnote{See also discussion in \citet{Sterelny-2014, Sterelny-2021}.}
            \end{quote}
    So, I also take the second conjunct of Premise (4) to be relatively uncontroversial, although I will expand upon this in more detail in section~\ref{sec:Main3}. In any case, Premise (5) follows logically from (3) and (4).

    Perhaps the most contentious of these Premises is (6). As was suggested in Section~\ref{sec:Language}, the primary purpose of linguistic communication is communication---i.e., information transfer. It is not the case that every feature of natural language is a requirement here. Nonetheless, the features of natural language that allow for robust information transfer are precisely the features that are understood to differentiate linguistic communication from simpler signalling systems---i.e., the compositional features of linguistic communication, among other things. These features give rise to systematicity and generalisability and, therefore, the flexibility and robustness of natural language for information transfer.

    There are actually two claims here. First, communication systems are best understood in terms of information transfer, and second, linguistic communication systems are uniquely robust. Even though the referent of `linguistic communication' is grossly underspecified, it is typically assumed that these capacities are unique to humans---namely, linguistic communication involves discrete, arbitrary, semantically meaningful signals whose referents may be displaced from the immediate environment \citep{LaCroix-2021}. In linguistic communication, these signals are {\it structured} and {\it flexible}. Although {\it proto-languages} can involve displaced reference and other semantic features of words, the word-like sequences thus produced are not syntactically organised \citep{Planer-Sterelny-2021}. Thus, researchers in linguistics, sociobiology, ethology, behavioural ecology, and related fields often take these claims for granted.%
            \footnote{See, e.g., \citet{Otte-1974, Green-Marler-1979, Seyfarth-et-al-1980a, Zahavi-1987, Hauser-1996, Smith-1997, Maynard-Smith-Harper-2003, Fitch-2008, Bradbury-Vehrencamp-2011} and discussion in \citet{LaCroix-2020}.}

    The conclusion, (7), is the {\bf main claim}, which follows logically from Premises (5) and (6). Hence, understanding value-alignment problems and language in the ways I have specified in Section~\ref{sec:Concepts} logically entails that linguistic communication is necessary for robust value alignment.
        \footnote{One may object that it is possible that some future technologies may allow for direct interface with human thought, thus implying linguistic communication is not uniquely robust---direct information transfer from the source would also be sufficiently robust. In this case, it would be possible to `read off' the principal's values directly, making it possible to align values without the need for linguistic communication. However, the {\it Language of Thought Hypothesis} (LOTH) proposes that thinking occurs in a `mental language', sometimes called {\it mentalese} \citep{Fodor-1975}. Assuming this is true, direct access to thought implies transferring information {\it via} a linguistic communication system, with the only distinction being modality. In the same way that it does not matter whether the transfer of information happens via written, spoken, or signed linguistic systems; it also does not matter whether the mode of linguistic communication is mental. Importantly, mentalese, like spoken, signed, or written natural language, is supposed to be compositional, implying that it has precisely the features I suggest are required for robust value alignment. The {\it Principle of Compositionality (of Mental Representations)} says that complex representations are composed of simple constituents, and the meaning of a complex representation depends upon the meanings of its constituents together with the constituency structure into which those constituents are arranged \citep{Rescorla-2019}. See also \citet{Fodor-1987, Fodor-2008, Normore-1990}.} %

\subsection{Objective Functions and Value Proxies}
    \label{sec:Main2}

    Another way to think about why language might plausibly be required for a robust form of value alignment is to consider how the {\it symbolic systems} approach to AI---sometimes called `good old-fashioned AI' (GOFAI)---is thought to have failed---or, perhaps more charitably, to be much more limited than initially believed.%
            \footnote{GOFAI is also called `first-wave AI' or `symbolic AI'. The phrase used here was coined by \citet{Haugeland-1985}. For discussions of the failures of GOFAI, see also \citet{McDermott-1987, Cantwell-Smith-2019}.} 
    Part of this is that these systems are far too rigid: every rule for action must be hard-coded. This is adequate for simple tasks, but as the complexity increases, it becomes intractable to write explicit instructions for every contingency.

    Part of why `second-wave' AI---ML and particularly DL methods bolstered by big data---has been surprisingly successful in comparison is that many rules for action are not hard-coded but implicit. What underwrites this intuition is that it is difficult to express the intuitive knowledge required for robust generalisation in the form of a set of verbally expressible rules that can be codified in a machine language. The symbolic systems approach of first-wave AI relied on the ability of humans to express explicit knowledge (often in the form of complicated {\it if-then} rules). In contrast, \citet{LaCroix-Bengio-2019} highlight that deep neural networks can `capture' precisely the kind of implicit knowledge that is difficult to express in a formal language.%
            \footnote{See also discussion in \citet{Buckner-2019}.}

    The failures of symbolic systems are supposed to have been caused by the system's being `brittle, unconducive to learning, defeated by uncertainty, and unable to cope with the world's rough and tumble' \citep{Cantwell-Smith-2019}. However, in contemporary approaches to AI, the `values' (objective functions) encoded in the systems are what drive learning. These objective functions are still hard-coded. Thus, although these systems are significantly more flexible for {\it learning} and {\it acting}, the values encoded in these systems are still brittle. By analogy, it appears that a similar move from rigidity to flexibility concerning these systems' objective functions will be necessary to ensure that the `values' of these systems are aligned with our values. In Section~\ref{sec:Main3}, I discuss, by analogy with humans, some empirical reasons why linguistic communication is necessary for robustly flexible values.

    One may object that inverse reinforcement learning (IRL) furnishes a counterexample to this claim. It does not. Whereas a classic RL model is given a reward function and attempts to learn behaviour based on the rewards it receives, an IRL model is given behaviour and attempts to learn the reward function that would give rise to that behaviour.%
            \footnote{Learning in this context may be through passive observation of behaviour \citep{Ng-Russell-2000, Abbeel-Ng-2004}, or it may involve active preference learning \citep{Markant-Gureckis-2014, Christiano-et-al-2017, Basu-et-al-2018}.} 
    Therefore, the objective function is learned instead of hard-coded. So, the thought goes, it may be possible for a sophisticated IRL model to learn values from behaviours insofar as actions transfer information. However, this line of reasoning is problematic in several directions.

    On the one hand, IRL is underspecified as a research problem because many reward functions can explain an actor's behaviour, and the costs of solving the problem tend to grow disproportionately with the size of the problem \citep{Arora-Doshi-2021}. Essentially, many different reward functions could explain any observed behaviour.%
        \footnote{This is effectively identical to the problem of rule-following, discussed in \citet{Wittgenstein-1953, Kripke-1982} and a significant body of secondary literature since.}

    On the other hand, behaviours are only ever going to serve as a proxy for values, which is part of why poorly specified objective functions give rise to value misalignment in the first place, as discussed in Section~\ref{sec:AIVAP}. Furthermore, by emphasising the information transferred by behaviours alone, we run into the well-known problems to which revealed preference theory from economics gives rise.%
        \footnote{Revealed preference theory \citep{Samuelson-1938a, Samuelson-1938b} uses consumer behaviour to analyse the choices made by individuals. See discussion in \citet{Sen-1973, Sen-1977, Sen-1993, Sen-1997, Sen-2002, Koszegi-Rabin-2007, Hausman-2012}.} 
    An IRL model's basic (and false) assumption is that the behaviour it observes {\it is} optimal, given the (hidden) reward function motivating that behaviour. In addition, even if behaviours were a good proxy for preferences and provided a high-fidelity channel for transferring information about those preferences, linguistic communication would aid learning objectives at a much higher rate. The rate will be highly relevant for certain aspects of {\it robust} value alignment. In addition to systematicity and generalisability, linguistic communication offers the possibility of speed to value alignment. In high-stakes cases, it will be impossible to wait until a sufficient number of behavioural instances are observed.

\subsection{The Curious Case of Human Value Alignment}
    \label{sec:Main3}

    As we have seen, the value-alignment problem for artificial intelligence asks how we design the {\it objective functions} of AI systems to guarantee maximal overlap between the actions of the artificial agent and the objectives or values of (human) principals. Despite the structural identity of value-alignment problems for AI systems and the principal-agent problem more generally, as I mentioned in Section~\ref{sec:AIVAP}, the solutions to these problems may be quite distinct. This point is made  clearer when we realise that principal-agent problems between humans allow for assumptions that are not warranted when the agent is an artificial system---namely, the ability to communicate one's {\it intended meaning} to an arbitrary degree of specificity via natural language.

    When considering problems in artificial intelligence, it is often instructive to ask, %
            {\it How do we do it?} 
    For the entire history of the species, humans have engaged in the production of agents with human-level intelligence, whose values may be misaligned with their own, and over which they have limited control---i.e., human children. This provides something of a {\it proof of concept} that values can be aligned in such agents \citep{Christian-2020}. From the cooperative nature of our species, it seems apparent that humans are at least capable of aligning their values in the variety of contexts we face daily. Part of the reason we can do so is that we can communicate via natural language. Humans use linguistic competence to impart extremely subtle norms, goals, and values in subsequent generations that align with a long cultural history of norms, goals, and values. Because human linguistic capacity comes hardwired, we can take this particular aspect of how we align our values for granted. Thus, the blind spot of value-aligned agency arises from a lack of sensitivity to the importance of language for {\it Homo sapiens} as fundamentally cooperative and social creatures.%
            \footnote{Of course, one might object that although humans use linguistic communication systems---i.e., they have linguistic communicative abilities---they often fail to align their values. This is fine. The {\bf main claim} furnished a {\it necessary} condition for value alignment; it does not say anything about whether language is {\it sufficient} for value alignment. It should be obvious that it is not. There is a substantive question about whether any set of abilities will suffice for value alignment; however, my claim is that no matter what set of abilities is uncovered, value alignment will be impossible without linguistic communication. The {\bf main claim} also does not imply linguistic communication is {\it the only} necessary condition for value alignment.}
    What is pressing in the context of value alignment for AI is that assumptions of linguistic competency cannot be taken for granted when considering artificial agents.

    Part of why language evolved in {\it Homo sapiens} is because of the cooperative demands that evolved in our lineage. In the context of hominin evolution, cooperation would have included `quite demanding forms of collective action; and with greater cooperation comes greater communication' \citep[73]{Planer-Sterelny-2021}. It is difficult to imagine how such robust cooperation could have evolved without corresponding increases in the complexity and flexibility of our communicative abilities. For example, the control of fire (it is supposed) would require communication systems much more complex than anything seen in other great ape species (though not necessarily language) because controlling fire poses a range of sophisticated coordination and cooperation problems.%
            \footnote{See, e.g., \citet{Ofek-2001, Gowlett-2006, Gowlett-2016, Garde-2009, Gowlett-Wrangham-2013, Pickering-2013, Gamble-2013} and discussion in \citet{Twomey-2013, Twomey-2014, Planer-Sterelny-2021}.}

    One can understand the intent of other humans precisely because one can communicate linguistically. %
            Strong empirical evidence exists for a tight, bidirectional connection between theory of mind capacities and linguistic capacities in human infants \citep{Astington-Jenkins-1999, Astington-Baird-2005, de-Villiers-2007, de-Villiers-de-Villiers-2014}. First, language appears necessary to learn to express concepts surrounding one's own feelings and inner world \citep{Nelson-2005, Dunn-Brophy-2005, Hutto-2008}; second, the information that language conveys about others' thoughts, feelings, beliefs, desires, etc., is much richer than the information conveyed through behaviour, eye gaze, gestural expressions, etc. \citep{Appleton-Reddy-1996, Harris-2005, Peterson-Siegal-1999, Wellman-Peterson-2013}; third, linguistic abilities allow individuals to reason abstractly about others' actions via their beliefs \citep{Astington-Jenkins-1999, de-Villiers-de-Villiers-2009, Milligan-et-al-2007}.

    Further, the compositional nature of language allows humans to communicate their goals with one another to an arbitrary degree of specificity.%
        \footnote{In addition, in the context fo AI, it has been suggested that linguistic communication may be necessary for transparency, trust, and explainability. For example, \citet{Williams-2018} argues that an ethical AI system that is asked for clarification may unintentionally communicate that it would be willing to perform an unethical action (even if it is unable to do so). Further, this type of miscommunication may negatively influence the morality of its human teammates. This point is discussed further in \citet{Jackson-2019-HRI, Jackson-Williams-2019-HRI}.} 
    Linguistic communicative abilities also affect the cultural accumulation of new concepts. High-fidelity cultural learning allows human populations to solve coordination/cooperation problems because it allows selective learning and the accumulation of small improvements over time \citep{Boyd-et-al-2011}. That is, language allows for accumulating knowledge across generations via social learning.%
            \footnote{See, e.g., \citet{Boyd-2016, Henrich-2016, Planer-Sterelny-2021}.} 
    \citet{Planer-Sterelny-2021} highlight that `some forms of cooperation are stable only if reputation (knowledge of the past social actions of others) is tracked reliably and is part of common knowledge' (25). Again, this robust sort of cooperation depends significantly upon language.

        Communication depends, to some extent, upon joint attention and common knowledge \citep{Tomasello-2008, Tomasello-2014}. Complex language is not necessary for joint commitment if there is common-ground understanding \citep{Tomasello-2014}; however, diminishing common ground between agents appears to necessitate greater lexical and structural richness in the language used to communicate information about disparate knowledge between agents.%
                \footnote{This been observed empirically in, e.g., children's sign languages; see \citet{Meir-et-al-2010}.}
        Essentially, when the social aspects of agent interactions become increasingly dispersed in time and space, members of a social group will need more sophisticated communication (and cognitive) abilities to report behaviour and events that happened `elsewhere and elsewhen' \citep[195]{Planer-Sterelny-2021}.

    Furthermore, human {\it moral} behaviour is rooted in uniquely-human linguistic abilities precisely because of the cultural accumulation that language `exemplifies and enables' \citep[19]{Bryson-2018}.%
        \footnote{See also discussion in \citet{Bryson-2008, Malle-Scheutz-2014, Malle-2015, Malle-2016}.} %
    \citet{Poulshock-2006-thesis} suggests that once complex language exists, humans have access to a new type of cultural evolution that can aid in promoting altruistic behaviour. A linguistically-encoded, abstract value system can promote and regulate moral behaviour to the extent that a sufficient number of members participating in such a moral system provides indirect benefits through, e.g., protection and group support. Language, therefore, allows humans to overcome genetic self-interest in a way not seen in other species in nature, despite a lack of direct benefits for moral behaviour. This relates to the importance of linguistic communication to delayed-return cooperation mentioned above (Section~\ref{sec:Main1}). Language allows social groups to encode moral principles and to demand adherence to those principles. It can turn high-cost behaviour into low-cost behaviour, thus aiding the enforcement of moral principles or norms \citep{Sober-Wilson-1998}. Furthermore, such moral codes can be transmitted across generations via spoken or written language.

    Language also provides a cost-efficient means for classifying certain behaviours as good or bad. It allows individuals to tag others as immoral (defectors/cheaters) or as moral (cooperators) via gossip, blame, praise, etc., when they adhere or fail to adhere to these principles; namely, linguistic communication allows us to articulate norms and monitor behaviour \citep{Boehm-2000}. Thus, linguistic ability significantly affects {\it moral} ability because it allows for a simple, efficient, systematic, generalisable, flexible, and (typically) high-fidelity means of transmitting information. %

    Empirical studies appear to corroborate the claim that moral behaviour depends (at least to some extent) on linguistic ability. For example, \citet{de-Waal-1996} suggests that the following (pre-)conditions are necessary for the evolution of morality: group value, mutual aid, and internal conflict. Actors must then resolve intragroup conflicts to balance individual and collective interests, either at the dyadic level (one-on-one interactions) or at a higher level. The latter may depend on community concern, which can be expressed in terms of mediated reconciliation or peaceful arbitration of disputes---as observed in many primate species, including chimpanzees. However, \citet{de-Waal-1996} notes that higher-level interactions may also be expressed in terms of indirect reciprocity (appreciation of altruistic behaviour at a group level) or encouragement of contributions to the quality of the social environment; importantly, these last two may be limited to {\it human} moral systems (34). But, it should be relatively obvious how language can support these functions---via, e.g., gossip, blame, praise, or encouragement.

    Considering the development of moral behaviour in humans, Several studies have suggested that moral judgements in children develop alongside language \citep{Smetana-Braeges-1990}. Infants appear to be prosocial, actively helping others from around 14 months of age \citep{Warneken-Tomasello-2006, Warneken-Tomasello-2007}.  \citet{Zahn-Waxler-et-al-1992} report that prosocial behaviours, such as sharing, providing comfort, etc., emerge in human children between years $1$ and $2$. These behaviours increase in frequency and variety over this period. However, Although human children may act in helpful ways and display empathic reactions by $2$ years, categorical judgements do not appear to emerge until around $3$ years of age.%
        \footnote{See discussion in \citet{Dahl-2018}, and studies in \citet{Zahn-Waxler-et-al-1992, Eisenberg-et-al-2007, Roth-Hanania-et-al-2011, Schmidt-et-al-2012, Smetana-et-al-2012, Dahl-Kim-2014, Dahl-2015}.}
    This is also around the time that, a `normative sense of obligation' begins to emerge \citep[250]{Tomasello-2018}. So, the distinction between moral and mere social norms, or conventions, appears to arise around three years.%
            \footnote{This distinction is more firmly established by four years, paralleling linguistic development in children. Early models of moral development in human children \citep{Piaget-1923, Piaget-1924, Piaget-1932, Kohlberg-1958, Kohlberg-1963, Kohlberg-1964, Kohlberg-1969, Kohlberg-1981, Kohlberg-1984} suggest a hierarchical development through several stages of understanding moral concepts---pre-conventional, conventional, and post-conventional---from pre-adolescence to adulthood. However, more recent studies have suggested that moral development occurs much earlier. It is also worth noting that the validity of these classical models has been called into question for several reasons, including possible gender bias \citep{Gilligan-1982}.}
    Importantly, for our purposes, these behaviours appear to develop and become more complex in parallel with linguistic development timelines.%
            \footnote{See extended discussion in \citet{Lust-2006}. Of course, progression rates will vary highly between individual children \citep{Brown-1973}.}
    Non-human primates, who lack linguistic capacities, display only {\it proto}-moral behaviours.%
            \footnote{There is little evidence of instrumental helping in chimpanzees, although there is a {\it tendency} for helping under certain conditions---for example, chimpanzees are more likely to help when the task does not involve food, when information about the goal can be transferred through a non-linguistic communication channel, and when the recipient of help is not a conspecific \citep{Warneken-Tomasello-2006}.}

    All these considerations are couched in the language of {\it morality}, which, like linguistic communication, is often considered uniquely human. One difficulty with such discussions is that it is not always clear what is meant by `morality' or what constitutes moral behaviour \citep{Dahl-2018}.%
        \footnote{See also discussion in \citet{Greene-2007, Wynn-Bloom-2014}.} 
    However, this is stronger than what is required for the {\bf main claim}. Aligning values is a type of coordination problem. \citet{Korsgaard-1996} suggests that the publicity of language allows values to be shared.%
        \footnote{In her view, {\it reasons} are essential to morality (which she approaches from a distinctly Kantian point of view). Part of this involves moral reflection from the first-personal perspective. In analogy with considerations concerning the impossibility of a private language \citep{Wittgenstein-1953}, Korsgaard thinks that there are no {\it private reasons}---reasons have a public normative force similar to how language has a public normative meaning. This view is somewhat complicated because it relies on Korsgaard's constructivist views surrounding identity; in any case, see discussion in \citep{Pauer-Studer-2003}, but cf. \citet{Gibbard-1999}.} 
    So, at a structural level, we can understand value alignment in terms of coordination and cooperation in addition to the problem of incentive structures. It is well understood that certain classes of coordination problems benefit from cheap talk \citep{Farrell-Rabin-1996}---i.e., simple communication. Therefore, simple communication channels will allow for simple alignment of values; however, robust (linguistic) communication is necessary to ensure robust value alignment in complex environments.

\phantom{a}

    To summarise, the first argument in favour of the {\bf main claim} depends upon the conceptualisation of language and value alignment, discussed in Section~\ref{sec:Concepts}. The {\bf main claim} follows logically from how these concepts are specified. The key constitutive features of linguistic communication---simplified here in terms of compositionality---give rise to the systematicity and generalisability that is a prerequisite for robust value alignment, understood as fundamentally a problem of information transfer. The second argument for the {\bf main claim} highlights that objective functions are rigid in precisely the ways that contributed to the failures of the symbolic systems approach to AI. Hence, although current systems are flexible regarding learning, they are still rigid regarding value alignment. Linguistic communication is necessary for flexible objective specification, which is necessary for robust value alignment. Finally, the connection between language and value alignment was explored in the context of cooperation in the hominin lineage. This provides additional empirical evidence of the necessity of language for robust value alignment while simultaneously highlighting the fact that we cannot take these capacities for granted when considering how to ensure that the values of AI systems are aligned with our own values.

    In the next section, I explore some consequences following from the {\bf main claim}.

\section{Discussion}
    \label{sec:Discussion}

    What follows if the {\bf main claim} is true? At minimum, the {\bf main claim} specifies a demanding lower bound on the difficulty of solving the value-alignment problem.

    At the same time, \citet{Bostrom-2014} suggests that achieving fully human-level performance on natural language is an `AI-complete' problem. This means that creating linguistic AI is essentially equivalent in difficulty or complexity to creating generally human-level AI systems (17). This implies that if one were to create a system fully capable of linguistic communication, in all likelihood, one would `also either already have succeeded in creating an AI that could do everything else that human intelligence can do, or they would be but a very short step from such a general capability' (17).%
            \footnote{Though, cf. \citet{Goebel-2008}.}

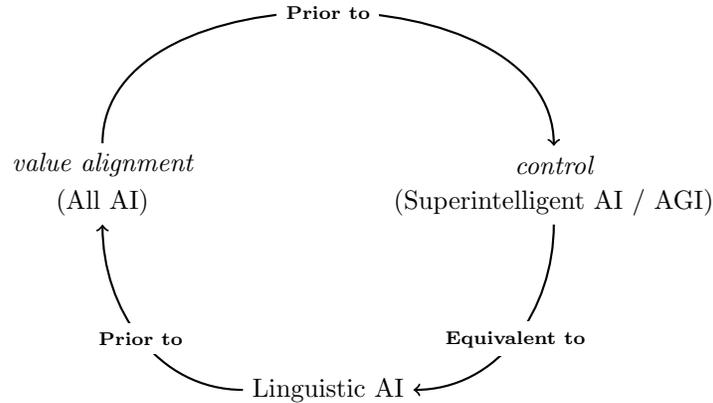
\begin{figure}[htb!]
        \centering
            \begin{tikzpicture}
                  \node (a) at (0,0) {{\it value alignment}};
                  \node (a1) at (0,-0.5) {(All AI)};
                  \node (b) at (6,0) {{\it control}};
                  \node (b1) at (6,-0.5) {(Superintelligent AI / AGI)};
                  \node (c) at (3,-3) {Linguistic AI};
                    \draw [->,thick] (a.north) to [out=90,in=90] node[midway, fill=white] {{\bf \scriptsize{Prior to}}} (b.north);
                    \draw [->,thick] (b1.south) to [out=270,in=0] node[midway, fill=white] {{\bf \scriptsize{Equivalent to}}} (c.east);
                    \draw [->,thick] (c.west) to [out=180,in=270] node[midway, fill=white] {{\bf \scriptsize{Prior to}}} (a1.south);
            \end{tikzpicture}
            \caption{Circularity following from the {\bf main claim} in conjunction with AI Completeness}
            \label{fig:Circularity}
        \end{figure}

    If \citet{Bostrom-2014} is correct about language being an AI-complete problem, then linguistic AI is functionally equivalent to human-level AI. If the {\bf main claim} is true, then linguistic AI is, in some sense, prior to robust value alignment. But, it was suggested above (\ref{sec:Control}) that value alignment is logically prior to problems of control.%
            \footnote{Recall that even narrow AI systems with limited functionality can fail to do what we intended, in the domain to which it is limited. In contrast, control problems arise in the context of stronger AI systems, on a par with human-level intelligence or superintelligence.}
    The conjunction of these three insights leads to circularity, which may imply that solving robust value alignment problems cannot be achieved until {\it after} we have already created an AI system that we are effectively unable to control; see Figure~\ref{fig:Circularity}. %

    This implication depends on a nested conditional: {\it If} language is an AI-complete problem, {\it then if} the {\bf main claim} is true, robust value alignment may be impossible---at least before the creation of an uncontrollable AI system. If any of the antecedents is false, then this may open the door to some (cautious) optimism about our ability to solve the value-alignment problem prior to the advent of a system that we cannot control. Fortunately, AI completeness is an {\it informal} concept, defined by {\it analogy} with complexity theory; so, it is possible that \citet{Bostrom-2014} is incorrect.%
            \footnote{Although, see discussion in \citet{Shahaf-Amir-2007}.}
    This leaves room for work on alignment with linguistic, albeit controllable, AI systems.

    In addition, the principal-agent model is highly idealised, as all models are. Hence, it may be that the {\bf main claim} depends upon an idealised (hence false) assumption. Still, it is useful for underscoring that informational asymmetries, rather than value misalignment {\it per se}, are the driving force of value-alignment problems. This insight opens the door to my argument that linguistic communication is at least a highly effective---if not necessary---means for transferring information to an arbitrary degree of specificity, which would allow on-the-spot adjustments in action in a way that is difficult to imagine accomplishing without language.%
            \footnote{Consider that any solution to a significantly complex value-alignment problem, driven by informational asymmetries, will require some degree of generalisability to accommodate unseen instances; and, that solution will require some degree of systematicity to accommodate efficiency constraints. As a result, such a solution will be inherently linguistic.}

    Counter-intuitively, the main driver of the value-alignment problem is {\it not} misaligned values but informational asymmetries. Thus, value-alignment problems are primarily problems of aligning {\it information}. However, this is not to say that values play no role in solving value-alignment problems. Information alone cannot {\it cause} action. Even in the classic economic and game-theoretic contexts, values are not objective. As \citet{Hausman-2012} highlights, a game {\it form}, complete with (objective) payoffs which correspond to states of the world conditional on actions, does not specify a {\it game} because it does not specify how the agents {\it value} the outcomes of the game form. For a game to be defined, it is not sufficient to know the objective states that would obtain under certain actions; instead, one must also know the {\it preferences} (values) of the actors over those (objective) outcomes. Thus, a specification of results needs to coincide with `how players understand the alternatives they face and their results' \citep[51]{Hausman-2012}.

    What comes out of the discussion of value alignment {\it qua} principal-agent problems is that the main issue involves aligning values by ensuring appropriate {\it incentive structures}. This is true in economics, because both the agent and principal are assumed to be utility maximisers. The principal must limit the divergence of the possible actions of the agent from the principal's interests---for example, by introducing incentives for the agent or paying some costs for monitoring to limit the agent's (possibly unaligned) actions. However, although principal-agent and value-alignment problems arise for the same reasons, there are dissimilarities in how they can be solved. For example, we can take certain things for granted when we consider interactions between human principals and human agents. In addition to linguistic communication abilities, human agents come with in-built values, whereas AI systems have their `values' programmed in an {\it objective function}. Given the complexity of accurately describing our objectives in a programming language, it is vanishingly unlikely that we can succeed in specifying the `correct' objective function for an artificial system---if such a specifiable function even exists---as was discussed in Section~\ref{sec:Main2}.

    One may object that `large language models' (LLMs)---e.g., BERT \citep{Devlin-et-al-2019}, GPT-3 \citep{GPT-3-2020}, RETRO \citep{Borgeaud-et-al-2022}, LaMDA \citep{Thoppilan-et-al-2022}, etc.---furnish a counterexample to the {\bf main claim} (and therefore its consequences). The thought is that LLMs {\it are} linguistic. However, this is misguided. The phrase `large language model' is a misnomer better described in terms of `large corpus models' \citep{Veres-2022} or `stochastic parrots' \citep{Bender-et-al-2021}. Some authors argue that meaning cannot be learned from form alone \citep{Bender-Koller-2020, Bisk-et-al-2020, Marcus-Davis-2020}; in this sense, LLMs do not capture meaning and, therefore, cannot be linguistic.%
        \footnote{See, e.g., \citet{McCoy-et-al-2019, Ettinger-2020, Pandia-et-al-2021, Sinha-et-al-2021, Sahlgren-Carlsson-2021}.}

    Although LLMs can be impressive syntax engines, language is more than mere syntax (Section~\ref{sec:Language}). Importantly, language is a primarily {\it social} endeavour, which requires the participation of both senders and receivers. Part of the reason that the output of LLMs {\it looks} so impressive is that it is being interpreted by competent language users (humans) who project meaning, intentions, understanding, etc., onto the system.

\begin{figure}[htb!]
            \centering
            \includegraphics[width=0.33\textwidth]{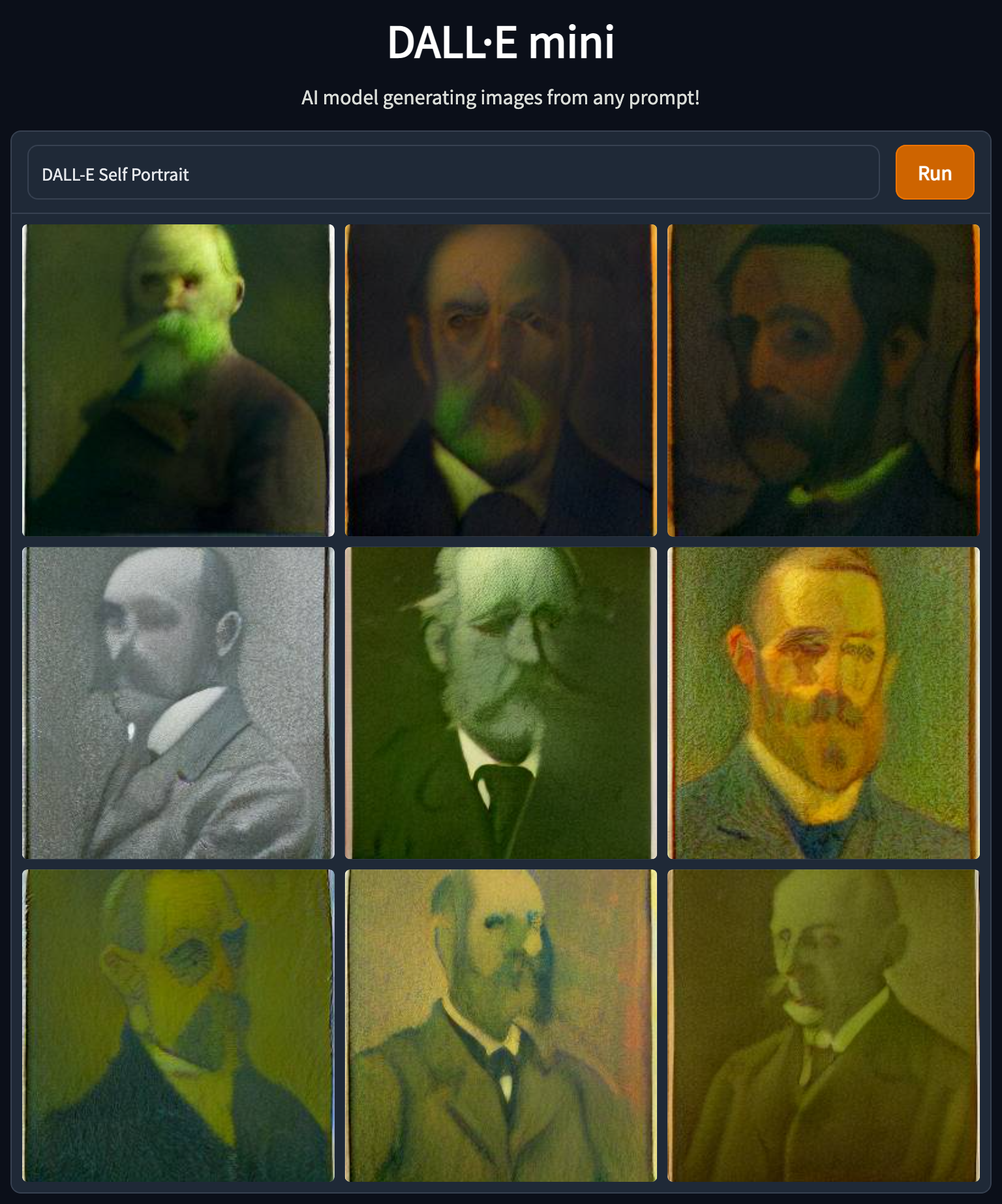}
            \caption{Output of DALL-E mini given prompt `DALL-E Self Portrait'. See \href{https://huggingface.co/spaces/dalle-mini/dalle-mini}{https://huggingface.co/spaces/dalle-mini/dalle-mini}}
            \label{fig:DallE}
        \end{figure}

    To illustrate this, prompting DALL-E---a version of GPT-3 that generates images from text inputs---to produce a `self-portrait' results in several images of apparently white, male humans; See figure~\ref{fig:DallE}.
    It would be strange to suggest from this that DALL-E `thinks' that {\it it} is a white, male human---a projection from the user's {\it interpretation} that a self-portrait applies (in fact) to {\it oneself}. What actually happens is this. DALL-E is trained on a dataset of text-image pairs. The output merely indicates that `self-portrait' is typically paired with images of white, male humans. There is nothing more to this than mere correlation, which is, at bottom, all that underlies present-day AI systems.

\section{Conclusion}
    \label{sec:Conclusion}

    The value-alignment problem is ubiquitous in the context of AI systems. As these systems become more sophisticated, these problems become more pressing. The {\bf main claim} helps clarify how difficult value alignment can be in this context. However, empirical evidence proves that aligning values between agents with human-level intelligence is at least {\it possible}. That said, it is important to maintain sensitivity to the informational asymmetries that underlie value alignment for AI systems since value alignment between human principals and human agents allow key features of the problem to be taken for granted---particularly, the ability to communicate linguistically. No such assumption can be made when the agent is artificial.

    It is precisely because we cannot make assumptions about, e.g., in-built linguistic ability that the implementation of ethically-aligned artificial systems comes to bear on normative theory. Normative theories historically focus on {\it human} agents. As such, they take for granted that agents can communicate linguistically. The creation of artificial systems allows for no such assumptions. As a result, thinking about problems of value alignment in terms of coordination---and linguistic aids to successful coordination---provides valuable insights into the implicit foundations upon which many normative theories rest.

    Thinking about the importance of linguistic communication for value alignment in artificial systems pushes work in this area beyond the straightforward application of specific insights from normative theory to AI by additionally providing novel insights in the opposite direction. Importantly, such insights will be theory-neutral insofar as they do not depend upon any particular metaethical framework.

    Hence, linguistic communication has been a blind spot in theorising about value alignment in the context of artificial agents; however, since the ability to communicate linguistically is taken for granted in normative theories that focus on human (natural) agents, little attention has been paid to the necessity of language for value-aligned agency in natural agents as well.

    As mentioned at the outset, I take the {\bf main claim} to hold in the context of value alignment more generally than just situations involving artificial agents. Hence, the linguistic blind spot of value-aligned agency applies in normative context for both natural and artificial agents.

\phantom{a}

\singlespacing
\paragraph{{\bf Acknowledgements}} I would like to thank the 
organisers of the Normative Theory and AI Workshop (2022)---Seth Lazar, Pamela Robinson, Claire Benn, and Todd Kharu---as well as participants---David Grant, Jeff Behrends, John Basl, Chad Lee-Stronach, Jack Parker, and David Danks---for helpful feedback. I would also like to thank, in no particular order,  
Aydin Mohseni, %
Jennifer Nagel, %
Yoshua Bengio, %
Dominic Martin, %
Benjamin Wald, %
Aaron Courville, %
Sasha Luccioni, %
Gillian Hadfield, %
Celso Neto, %
Duncan McIntosh, %
Aaron Wright, %
Jeffrey Barrett, %
 Simon Huttegger, 
 Richmond Campbell, 
and Michael Noukhovitch, %
for discussion and feedback throughout the process. This paper was presented at the Philosophy Department Colloquium at Dalhousie University (2021). I am grateful to the audience members for helpful discussion. %
Thanks also to the Schwartz Reisman Institute at the University of Toronto for partially funding this research, and to Mila - Qu{\'e}bec Artificial Intelligence Institute for providing generous resources.

\newpage
\singlespacing
\bibliographystyle{apalikelike}
\bibliography{Biblio}
\end{document}